\documentclass[10pt, conference]{ieeeconf}
\IEEEoverridecommandlockouts 

\usepackage{svg}
\usepackage{graphicx}
\usepackage{xcolor}
\usepackage{tikz}
\usepackage{amsmath,amssymb}
\usetikzlibrary{arrows.meta}
\usepackage{pgfplots,amsmath,amssymb}
\usepackage{pdfpages}
\usepackage{algorithmic}
\usepackage{algorithm}
\usepackage{array}
\usepackage{booktabs}
\usepackage{multirow}
\usepackage[caption=false,font=normalsize,labelfont=sf,textfont=sf]{subfig}
\usepackage{caption}
\usepackage{textcomp}
\usepackage{stfloats}
\usepackage{url}
\usepackage{verbatim}
\usepackage{graphicx}
\usepackage{cite}
\usepackage[hidelinks]{hyperref}
\usepackage{cleveref}
\usepackage{xcolor}
\usepackage{cleveref}
\usepackage{cuted}

\usepackage{amssymb,tikz}
\usetikzlibrary{arrows.meta}

\newcommand{\ie}{\textit{i.e.,}}%
\newcommand{\eg}{\textit{e.g.,}}%
\newcommand{\etal}{\textit{et al.}}%
\newcommand{\ci}[1]{{\scriptsize $[#1]$}}
\definecolor{todoColor}{HTML}{FF3B30} 
\definecolor{icColor}{HTML}{007AFF}   
\definecolor{jColor}{HTML}{00A676}    
\definecolor{wColor}{HTML}{AF52DE}    
\definecolor{cColor}{HTML}{FF9500}    

\begin{document}

\title{
Hierarchical Aggregation of Semantic Uncertainty in 3D Scene Graphs
}

\author{
    Carlos Cueto Zumaya\authorrefmark{2},
    Iacopo Catalano\authorrefmark{2},
    Wallace Moreira Bessa,
    and Julio A. Placed%
    \thanks{\authorrefmark{2} Equal Contribution. Iacopo Catalano, Carlos Cueto Zumaya, Wallace Moreira Bessa are with the University of Turku, Finland. Julio A. Placed is with the Arag\'on Institute of Technology (ITA) and the University of Zaragoza, Spain. This work was partially supported by the Finnish Cultural Foundation, the Finnish Ministry of Education and Culture through the Intelligent Work Machines Doctoral Education Pilot Program (IWM VN/3137/2024-OKM-4), and by DGA\_FSE T73\_23R. Corresponding: \texttt{\small crcuzu@utu.fi}}
}

\maketitle

\preCutedStrip={\vskip-60pt}
\begin{strip}
    \centering
    \includegraphics[
        width=0.9\textwidth
    ]{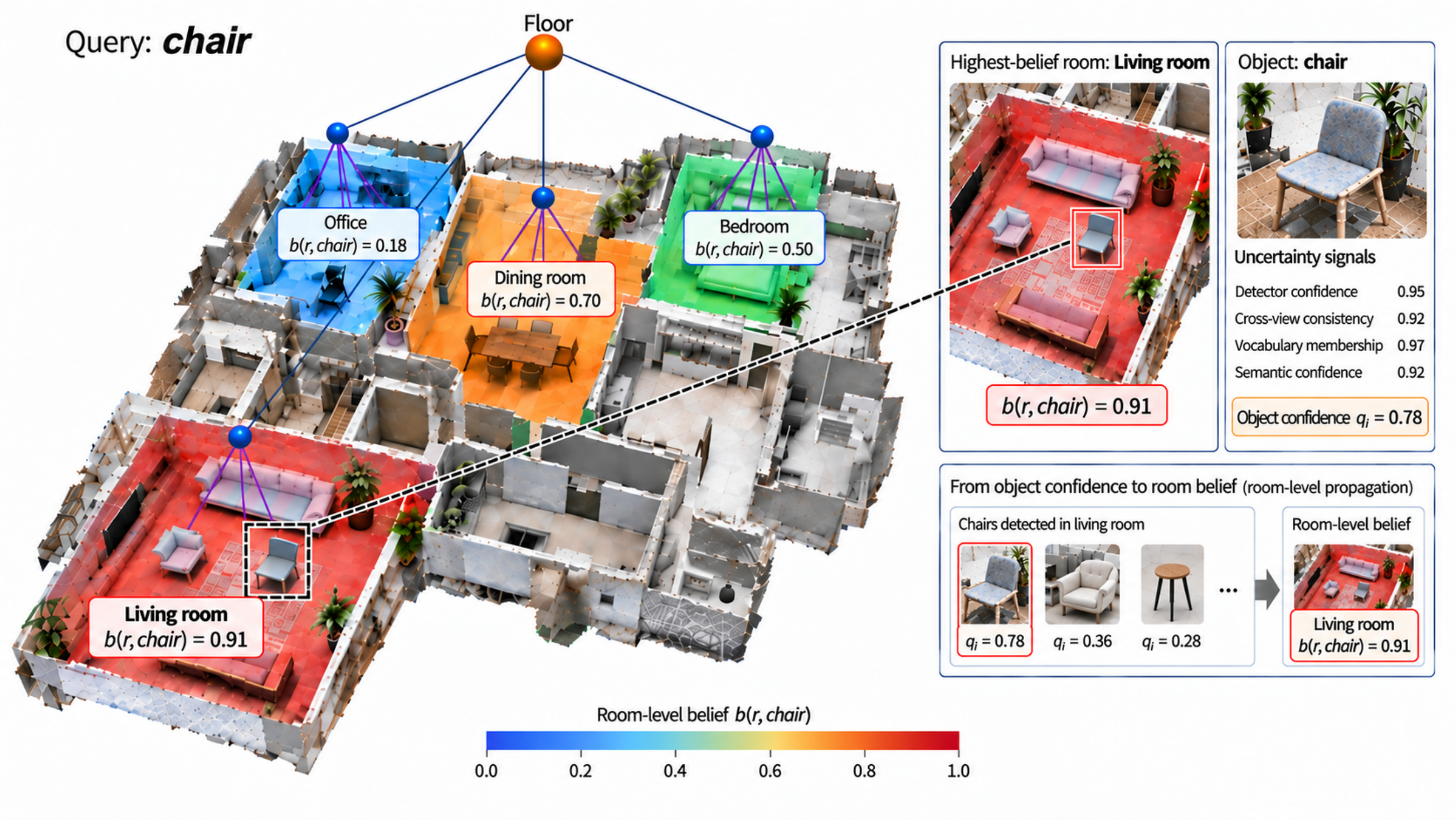}
    \captionof{figure}{\textbf{Visualization of uncertainty propagation from objects to rooms.}
    For the query class \emph{chair}, each room in the 3D scene graph is colored according to its propagated room-level belief $b(r,\mathrm{chair})$, with higher values indicating stronger evidence that the queried class is present in the room. The inset illustrates the object-level evidence contributing to this belief for a chair in the highest-belief room: detector confidence, cross-view consistency, vocabulary-membership probability, and semantic confidence are combined into the object reliability $q^{}_i$. This object-level uncertainty is propagated through the scene-graph hierarchy to obtain the corresponding room-level beliefs.}
    \label{fig:belief-teaser}
\end{strip}

\begin{abstract}
    Open-vocabulary 3D Scene Graphs (3DSGs) ground each object node in a vision-language embedding, yet they record every entry as equally certain, so a robot querying the map cannot tell which of its entries are unreliable. Estimators of semantic uncertainty could supply that distinction, but they require repeated sampling of a model, training, or held-out labels, none of which are available to a deployed system at query time. We present a framework that exploits the detector confidence and the embeddings a 3DSG already stores, converts them into a probability that an entry is correct, and propagates that probability through the containment hierarchy into a belief that a room contains a queried class. Four signals, each paired with the object-level error it indicates, are converted to probabilities at the logit scale learned by the vision-language model and combined in closed form with no additional perception or training. Objects sharing a detector and a vocabulary fail together, so the framework aggregates them in the fully correlated limit, where an aggregation under independence would treat one repeated error as repeated evidence. Evaluated on HM3DSem against a state-of-the-art 3DSG system, the framework improves object retrieval and lowers the error of the room-level assertions of the graph it reads.
\end{abstract}

\section{Introduction}\label{sec:introduction}

Semantic 3D mapping builds 3D representations of an environment annotated with semantic information extracted by visual perception~\cite{mccormac2017semanticfusion}. The resulting maps are systematically overconfident: each entry is fused from noisy, correlated observations integrated as if independent, and the map records every entry as equally certain~\cite{marques2024overconfidence}. A robot answering queries (\eg{} which rooms contain a chair, where the nearest sink is) has no indication of which entries are unreliable.

3D Scene Graphs (3DSGs) organize entries into a hierarchy of increasing abstraction, with containment relationships between nodes of different layers (\eg{} object nodes are contained in room nodes, rooms in floors)~\cite{catalano20253d}. Open-vocabulary 3DSG pipelines adopt vision and language models (VLMs/LLMs) as active grounding mechanisms, assigning object nodes with open-vocabulary semantic descriptions~\cite{deng2024opengraph, werby2024hierarchical, puigjaner2026relationship}. During graph construction, these systems compute confidence signals of each detection and the visual and text embeddings underlying each assignment. However, at query time these are discarded, and the graph is read as if every node were certain.

This paper shows that the quantities the graph already stores are sufficient to define per-object uncertainty signals and to propagate them, as probabilities, through the hierarchy, yielding room-level beliefs of the form ``this room contains a chair with probability $p$'' that are more discriminative and better calibrated than the raw outputs of the same systems (\cref{fig:belief-teaser}). The robot thereby maintains, alongside the map, an estimate of the reliability of each entry, structured by the topology of the scene. Unlike the covariances that pose-graph optimization maintains over geometry~\cite{millan2026generation}, the uncertainty we quantify concerns what the map asserts about the world (labels, presence, containment), and it is consumed by task-level decisions rather than by the state estimator.

The closest prior work~\cite{zhang2026remember} computes semantic uncertainty but requires additional model queries for every object, and it remains at the object level.

The main contributions of this work are:
\begin{itemize}
    \item \textbf{Per-object semantic uncertainty from quantities the graph already stores.} We define four signals over the detector confidence and the embeddings of an open-vocabulary 3DSG and convert them to probabilities at the logit scale of the vision-language model itself, without additional model queries, training, or held-out labels.

    \item \textbf{A correlation-aware aggregation over the containment hierarchy.} We factorize the per-object probability by the chain rule, show the product of the four signals to be a conservative estimate of it, bracket the room-level belief between its independent and its fully correlated limits, and adopt the fully correlated limit.
\end{itemize}
\section{Related Work}
\label{sec:related_work}

We review prior work along the two axes that define our contribution: whether an uncertainty signal is computable from what the mapping system stores or requires further model queries, training, or held-out data; and the level at which it is quantified (the object or the room a planner queries). Two bodies of work are relevant: the 3D scene graphs that expose the quantities we read, and the quantification of semantic uncertainty.

\subsection{3D Scene Graphs}

Hierarchical 3DSGs represent an environment as a layered graph linking geometry, semantics, and topology~\cite{catalano20253d}. 
Early pipelines~\cite{armeni20193d} relied on segmentation and detection methods that are class specific, but were limited to a predefined set of concepts at inferences, limiting their applicability to real-world robotics. Open-vocabulary variants enhance the grounding mechanism via VLMs and LLMs, assigning to nodes open-vocabulary semantic descriptions obtained from class-agnostic segmentation methods. Common strategies include assigning object labels by comparing a visual embedding of each object against text embeddings of candidate labels~\cite{werby2024hierarchical, puigjaner2026relationship}, or annotating objects with captions generated by a VLM while also storing language-aligned embeddings~\cite{nyffeler2025hierarchical, gorlo2026describe}. Despite their strategies, none of them quantifies the uncertainty of these assignments.

Recent works~\cite{saucedo2024belief, saucedo2025estimating} predict likely unobserved objects from commonsense priors to guide object search, adding beliefs to a 3DSG. However, their belief concerns content the map does not yet contain, and it is neither calibrated nor propagated through the containment hierarchy.

Closest to our work, Zhang~\etal~\cite{zhang2026remember} attaches a semantic uncertainty score to the objects of a 3DSG by querying a VLM for several captions of each object across viewpoints and scoring each object by the dispersion of those captions. It acquires further views for the objects of highest dispersion. Their score requires additional model queries for every object, and it remains at the object level; its evaluation measures uncertainty reduction and downstream accuracy, and does not assess the calibration of the score.

None of these systems quantifies the reliability of its stored content or propagates that reliability through the containment hierarchy. We treat the detection confidence, the visual and text embeddings, and the containment hierarchy as the interface to this class of systems.

\subsection{Quantifying Semantic Uncertainty}

In an open-vocabulary map, each label is produced by a vision-language model, so the reliability of a label is the reliability of a model output. On the models themselves, that reliability is estimated by sampling the model repeatedly and measuring how far the responses disagree; wide dispersion among the samples indicates an unreliable or hallucinated output~\cite{farquhar2024detecting}. The same principle applies to VLMs, where the estimate combines the dispersion of the sampled responses with a learned ambiguity term~\cite{lau2026uncertainty, zhang2026fuse}. Each such estimate requires several model queries per object, and the learned variants require training data.

The same reliability is available from a single forward pass. Matching an input against a fixed vocabulary produces a profile of similarities whose shape already reflects it: the mass concentrates on one class for a confident match and spreads near-uniformly when the input fits no class~\cite{ming2022delving}. A set of negative references, terms selected once for their distance from every class in the vocabulary, provides an explicit comparison that needs no training and no repeated queries~\cite{jiang2024negative}.

These estimators quantify the reliability of a single input in isolation. A 3D scene graph instead stores one fused embedding per object, links objects to the rooms that contain them, and is queried at the room level. We read the single-pass estimate from the embeddings the map already holds, without extra queries or training, and combine the per-object results along the containment edges into a belief about the room. Objects labeled by one model against one vocabulary fail together rather than independently, and we account for this dependence when aggregating.
\section{Methodology}
\label{sec:methodology}

\begin{figure*}[t]
    \centering
\definecolor{sigcol}{HTML}{009E73}   
\definecolor{outcol}{HTML}{0072B2}   
\definecolor{roomcol}{HTML}{E69F00}  
\definecolor{inkcol}{HTML}{3D3D3A}

\providecommand{\Ed}{E^{}_{s}}
\providecommand{\Ev}{E^{}_{\mathrm{view}}}
\providecommand{\Em}{E^{}_{\mathrm{mem}}}
\providecommand{\Esm}{E^{}_{\mathrm{sem}}}

\providecommand{\probbar}[4]{%
  \fill[gray!22,rounded corners=0.6pt] (#1,#2) rectangle ++(1.15,0.22);
  \fill[#4,rounded corners=0.6pt] (#1,#2) rectangle ++({1.15*#3},0.22);
  \draw[gray!50,rounded corners=0.6pt,line width=0.25pt] (#1,#2) rectangle ++(1.15,0.22);
  \node[font=\scriptsize,anchor=west,text=gray!55!black] at (#1+1.25,#2+0.11) {#3};
}
\providecommand{\bigarrow}[3]{%
  \draw[-{Latex[length=9pt,width=10pt]},black!42,line width=4.5pt,
        line cap=butt] (#1,#3) -- (#2,#3);
}

\begin{tikzpicture}[x=1cm,y=1cm]

\node[font=\scriptsize\bfseries,text=gray!35!black,anchor=south] at (1.75,6.10)
      {3D scene graph (\S~\ref{sec:scene-graph})};

\node[draw=roomcol!70,fill=roomcol!12,rounded corners=2pt,line width=0.5pt,
      minimum width=2.4cm,minimum height=0.48cm,font=\footnotesize,
      text=inkcol] (room) at (1.75,5.61) {room $r$};

\foreach \x/\n in {0.85/1, 2.65/3}{
  \node[circle,draw=gray!55,fill=gray!10,line width=0.4pt,inner sep=0pt,
        minimum size=0.50cm,font=\scriptsize,text=inkcol] (o\n) at (\x,4.55) {$o^{}_{\n}$};
    \draw[gray!55,line width=0.4pt] (room.south) to[out=-90,in=100] (o\n.north);
}
\node[circle,draw=sigcol!70,fill=sigcol!14,line width=0.7pt,inner sep=0pt,
      minimum size=0.50cm,font=\scriptsize,text=inkcol] (oi) at (1.75,4.55) {$o^{}_i$};
\draw[sigcol!60,line width=0.5pt] (room.south) -- (oi.north);

\draw[gray!55,line width=0.35pt,dash pattern=on 1.5pt off 1.5pt]
      (oi.south) -- (1.75,3.92);

\fill[gray!4,rounded corners=2pt] (0.30,1.70) rectangle (3.0,3.90);
\draw[gray!45,rounded corners=2pt,line width=0.35pt] (0.30,1.70) rectangle (3.0,3.90);
\node[anchor=west,font=\scriptsize\itshape,text=gray!55!black] at (0.48,3.66)
      {Interface~\eqref{eq:interface}};
\foreach \y/\who/\qty in {%
  3.10/{Per object}/{$s^{}_i,\ \mathbf{v}^{}_i$},
  2.55/{Per edge}/{$r(\cdot)$},
  2.00/{Per graph}/{$\{\mathbf{t}^{}_c\}_{c\in\mathcal{C}}$}}{
  \node[anchor=west,font=\scriptsize,text=gray!62!black] at (0.48,\y) {\who};
  \node[anchor=west,font=\scriptsize,text=inkcol] at (1.82,\y) {\qty};
}

\node[font=\scriptsize\bfseries,text=gray!35!black,anchor=south] at (8.48,6.10)
      {Per-object signals, in the conditioning order of eq.~\eqref{eq:obj-prob-chain} (\S~\ref{sec:uncertainty-signals})};
 
\foreach \y/\lbl/\val/\err in {%
  5.2500/{$s^{}_i \approx P(\Ed)$}/0.90/{Detection confidence {\scriptsize\color{gray!62!black}$\cdot$ false positive}},
  4.1267/{$P^{\mathrm{view}}_i \approx P(\Ev \mid \Ed)$}/0.95/{Cross-view agreement {\scriptsize\color{gray!62!black}$\cdot$ fusion error}},
  3.0033/{$P^{\mathrm{mem}}_i \approx P(\Em \mid \Ed,\Ev)$}/0.80/{Membership in $\mathcal{C}$ against negatives $\mathcal{N}$ {\scriptsize\color{gray!62!black}$\cdot$ unknown class}},
  1.8800/{$\bar{P}^{\mathrm{sem}}_i \approx P(\Esm \mid \Ed,\Ev,\Em)$}/0.70/{Semantic and coherence margins {\scriptsize\color{gray!62!black}$\cdot$ labeling error}}}{
  \fill[sigcol!7,rounded corners=2pt] (4.10,\y-0.18) rectangle (12.85,\y+0.60);
  \draw[sigcol!45,rounded corners=2pt,line width=0.35pt] (4.10,\y-0.18) rectangle (12.85,\y+0.60);
  \node[anchor=west,font=\footnotesize,text=inkcol] at (4.32,\y+0.40) {\err};
  \node[anchor=west,font=\footnotesize,text=inkcol] at (4.32,\y+0.02) {\lbl};
  \probbar{10.60}{\y+0.11}{\val}{sigcol!55}
}
\foreach \y in {5.2500,4.1267,3.0033}{
  \draw[-{Latex[length=3.2pt,width=2.4pt]},sigcol!65,line width=0.5pt]
        (8.48,\y-0.20) -- (8.48,\y-0.50);
}

\node[font=\scriptsize\bfseries,text=gray!35!black,anchor=south] at (15.60,6.10)
      {Hierarchical aggregation (\S~\ref{sec:propagation})};

\fill[outcol!8,rounded corners=2pt] (13.75,4.75) rectangle (17.45,5.85);
\draw[outcol!55,rounded corners=2pt,line width=0.35pt] (13.75,4.75) rectangle (17.45,5.85);
\node[anchor=west,font=\footnotesize,text=inkcol] at (13.93,5.66)
      {Object probability {\scriptsize\color{gray!62!black}$\cdot$ Eq.~\eqref{eq:obj-prob}}};
\node[anchor=west,font=\footnotesize,text=inkcol] at (14.53,5.19) {$q^{}_i$};
\probbar{15.15}{5.09}{0.48}{outcol!60}

\draw[-{Latex[length=4pt,width=3pt]},gray!65,line width=0.5pt] (15.60,4.73) -- (15.60,4.40);

\fill[roomcol!6,rounded corners=2pt] (13.75,1.70) rectangle (17.45,4.40);
\draw[roomcol!60,rounded corners=2pt,line width=0.5pt] (13.75,1.70) rectangle (17.45,4.40);
\node[anchor=west,font=\scriptsize,text=gray!62!black] at (13.93,4.22)
      {Room $r$: objects labeled $c$};
\foreach \x/\v in {13.93/0.48, 15.00/0.31, 16.07/0.22}{
  \fill[outcol!55,rounded corners=1.5pt,opacity={0.20+\v}] (\x,3.40) rectangle ++(0.95,0.60);
  \draw[outcol!50,rounded corners=1.5pt,line width=0.35pt] (\x,3.40) rectangle ++(0.95,0.60);
  \node[font=\scriptsize,text=inkcol] at (\x+0.475,3.70) {\v};
}
\draw[roomcol!85,rounded corners=1.5pt,line width=1pt] (13.93,3.40) rectangle ++(0.95,0.60);

\node[anchor=west,font=\footnotesize,text=inkcol] at (13.93,2.92)
      {Room belief {\scriptsize\color{gray!62!black}$\cdot$ Eq.~\eqref{eq:room-belief}}};
\node[anchor=west,font=\scriptsize,text=inkcol] at (13.93,2.46) {$b(r,c)$};
\probbar{15.15}{2.35}{0.48}{roomcol!70}
\node[anchor=west,font=\scriptsize,text=gray!62!black] at (13.93,1.98) {noisy-OR};
\probbar{15.15}{1.87}{0.72}{gray!40}

\bigarrow{3.23}{3.90}{3.98}
\bigarrow{13.0}{13.70}{3.98}

\end{tikzpicture}
    \caption{\textbf{Overview of the framework.} The 3DSG supplies the interface of \cref{eq:interface}; four signals are computed from it in the conditioning order of \cref{eq:obj-prob-chain}, each paired with the error type it indicates, and multiplied into the per-object probability $q^{}_i$ of \cref{eq:obj-prob}. Over the objects of room $r$ labeled $c$, the belief of \cref{eq:room-belief} is the fully correlated limit of the bracket of \cref{eq:bracket}, shown against the aggregation under independence. The negative set $\mathcal{N}$ is mined once from a lexical database. Values are illustrative.}
    \label{fig:method}
\end{figure*}
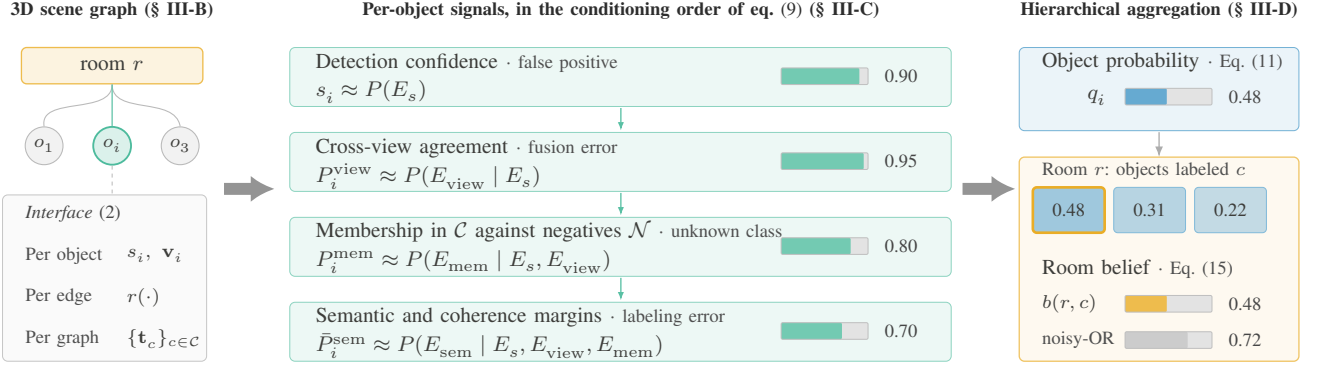

\subsection{Problem Formulation}
\label{sec:problem}

We consider a robot equipped with a semantic map $\mathcal{M}$ of an environment, containing a set of object entries $\mathcal{O} \subseteq \mathcal{M}$ and a set of rooms $\mathcal{R} \subseteq \mathcal{M}$. Each object entry $o^{}_i \in \mathcal{O}$ carries a label $\ell^{}_i \in \mathcal{C}$ from a set of classes $\mathcal{C}$, a position $\mathbf{x}^{}_i \in \mathbb{R}^3$, and an assignment to a room $r(o^{}_i) \in \mathcal{R}$.

Let $y(r,c) \in \{0,1\}$ denote the true presence of an object of class $c$ in room $r$. For a room $r \in \mathcal{R}$ and a class $c \in \mathcal{C}$, we estimate a belief
\begin{equation}
    b(r, c) \approx P\big(y(r,c) = 1 \mid \mathcal{M}\big) \in [0, 1]
    \label{eq:belief}
\end{equation}
that room $r$ contains at least one object of class $c$. The belief reads the entries $\mathcal{M}$ holds, so an instance the pipeline never reconstructed lies outside its domain: for such a pair $b(r,c) = 0$ while $y(r,c) = 1$. The frequency of the event measures the recall of the construction pipeline, and is reported separately.

The belief must be \emph{discriminative} (it must assign higher values where the class is present than where it is absent) and \emph{calibrated} (among the room-class pairs assigned belief close to $p$, the fraction that contain the class must be close to $p$). A valid belief is a function of the contents of $\mathcal{M}$ and of scene-independent constants derived from label text alone.

\subsection{Hierarchical 3D Scene Graph}
\label{sec:scene-graph}

A 3DSG is a map $\mathcal{M}$ in the form of a layered graph whose nodes represent entities at increasing spatial scale (\eg{} objects $\mathcal{O}$ and rooms $\mathcal{R}$) and whose edges encode containment between them~\cite{catalano20253d} (the containment edge of each object $o^{}_i \in \mathcal{O}$ fixes its room assignment $r(o^{}_i) \in \mathcal{R}$).

An open-vocabulary graph~\cite{werby2024hierarchical, puigjaner2026relationship, gorlo2026describe} labels each object $o^{}_i$ by comparing a visual embedding $\mathbf{v}^{}_i$ of the object, fused from its observations across views, against a form  of text or caption embeddings $\{\mathbf{t}^{}_c\}^{}_{c \in \mathcal{C}}$, assigning $\ell^{}_i = \arg\max_{c} \cos(\mathbf{v}^{}_i, \mathbf{t}^{}_c)$ (the label of the class of greatest alignment with $\mathbf{v}^{}_i$). Its construction produces two further per-object quantities, stored at negligible cost: the detector confidence $s^{}_i \in [0,1]$, and the visual and text embeddings underlying the assignment.

The interface is therefore three quantities: the detector confidence ($s^{}_i$), a language-aligned visual embedding per object ($\mathbf{v}^{}_i,\, \{\mathbf{t}^{}_c\}_{c \in \mathcal{C}}$), and the containment hierarchy ($r(\cdot)$). Formally, we define the interface tuple
\begin{equation}
    \mathcal{I} = \big(s^{}_i,\, \mathbf{v}^{}_i,\, \{\mathbf{t}^{}_c\}_{c \in \mathcal{C}},\, r(\cdot)\big),
    \label{eq:interface}
\end{equation}
and any 3D scene graph that exposes $\mathcal{I}$ supports our method.

\subsection{Uncertainty Signals}
\label{sec:uncertainty-signals}

From the quantities of $\mathcal{I}$ we derive four per-object uncertainty signals, each paired with the object-level error it indicates.
\Cref{fig:method} summarizes the signals
and their aggregation into the room-level belief.

\subsubsection{Detection Confidence}
The detector reports a confidence score $s^{}_i \in [0,1]$ for the detection. This score is produced differently across systems,~\eg{} from SAM's~\cite{kirillov2023segment} predicted IoU~\cite{werby2024hierarchical}, GroundingDINO's~\cite{liu2023grounding} detection score~\cite{ zumaya2026occupancy}, or YOLOE's~\cite{wang2025yoloe} predicted IoU~\cite{puigjaner2026relationship}. In each case, the score is correlated with detection quality but not guaranteed to be calibrated; we treat $s^{}_i$ as an estimate of the probability that the detection corresponds to a real object, consistent with common practice for detector confidence scores. The associated error is a false positive: a detection that corresponds to no real object.

\subsubsection{Semantic Uncertainty}
The label $\ell^{}_i$ is the class most similar to $\mathbf{v}^{}_i$, and we quantify its reliability as the amount by which that similarity exceeds the similarity of the next class. Image and text embeddings occupy two separate regions of the joint space, a gap known as the modality gap~\cite{liang2022mind}. As a consequence, the absolute value of $\cos(\mathbf{v}^{}_i, \mathbf{t}^{}_{\ell_i})$ is shaped by this gap as much as by the correctness of $\ell^{}_i$, and is neither comparable across objects nor interpretable as a probability on its own. A near-synonym of $\ell^{}_i$ (\eg{} sofa for couch) has an almost identical text embedding, and the difference between the 
similarities would then be small for reasons of vocabulary rather than for properties of the object. We therefore measure the similarity of $\ell^{}_i$ against the highest-scoring class whose text embedding is not near-identical to $\mathbf{t}^{}_{\ell_i}$,
\begin{equation}
    m^{}_i = \cos\bigl(\mathbf{v}^{}_i, \mathbf{t}^{}_{\ell_i}\bigr) - \max_{c\, :\, \cos(\mathbf{t}^{}_c,\, \mathbf{t}^{}_{\ell_i}) < \tau} \cos\bigl(\mathbf{v}^{}_i, \mathbf{t}^{}_c\bigr),
\end{equation}
where $\tau < 1$ is a similarity threshold. The constraint removes from the maximum every label synonymous with $\ell^{}_i$ and, since $\cos(\mathbf{t}^{}_{\ell_i},\, \mathbf{t}^{}_{\ell_i}) = 1$, the label $\ell^{}_i$ itself, so that $m^{}_i$ compares $\ell^{}_i$ against the best genuinely distinct alternative. At CLIP's~\cite{radford2021learning} logit scale $\alpha$, the model's own softmax assigns each class $c$ a logit $\alpha\cos(\mathbf{v}^{}_i, \mathbf{t}^{}_c)$. Restricting this softmax to the binary comparison between $\ell^{}_i$ and the next best distinct class gives the model's confidence that $\ell^{}_i$ is the correct class rather than the nearest distinct one,
\begin{equation}
    \begin{aligned}
        P^{\mathrm{sem}}_i &= \frac{e^{\alpha\cos(\mathbf{v}^{}_i, \mathbf{t}^{}_{\ell_i})}}{e^{\alpha\cos(\mathbf{v}^{}_i, \mathbf{t}^{}_{\ell_i})} + e^{\alpha \max_{c\,:\,\cos(\mathbf{t}^{}_c,\, \mathbf{t}^{}_{\ell_i}) < \tau}\cos(\mathbf{v}^{}_i, \mathbf{t}^{}_c)}} \\ &= \sigma\bigl(\alpha\, m^{}_i\bigr),
    \end{aligned}
    \label{eq:semantic}
\end{equation}
with $\sigma$ the logistic function, a two-class softmax reducing algebraically to a sigmoid of the logit gap. The semantic uncertainty is its complement $1 - P^{\mathrm{sem}}_i$. Restricting the denominator to two classes removes the remaining $|\mathcal{C}| - 2$ terms, so $P^{\mathrm{sem}}_i$ is an upper bound on the probability the full softmax over $\mathcal{C}$ assigns to $\ell^{}_i$; the quantity it estimates is the correctness of $\ell^{}_i$ against the nearest distinct alternative, a label synonymous with $\ell^{}_i$ counting as correct.
The associated error is a labeling error: an assigned class that does not match the object.

\subsubsection{Label Coherence}
The semantic margin $m^{}_i$ of \cref{eq:semantic} measures $\mathbf{v}^{}_i$ against text references, which requires crossing the offset that separates image from text embeddings~\cite{liang2022mind}. Comparisons within one modality (\eg{} image--image) do not cross that offset, so their level is comparable across objects. For each class $c$ instantiated in $\mathcal{M}$, let $\boldsymbol{\mu}^{}_c$ be the normalized mean of the visual embeddings of the objects labeled $c$, an in-map visual reference, or \emph{prototype}, for the class. We define the coherence margin $m'_i$ by comparing the similarity of $o^{}_i$ to its label's prototype against that of the next best distinct prototype. When evaluating $o^{}_i$ we omit its embedding $\mathbf{v}^{}_i$ from the mean defining $\boldsymbol{\mu}^{}_{\ell_i}$. If $\mathbf{v}^{}_i$ were included, $\boldsymbol{\mu}^{}_{\ell_i}$ would be computed partly from $\mathbf{v}^{}_i$ itself, and $\cos(\mathbf{v}^{}_i, \boldsymbol{\mu}^{}_{\ell_i})$ would then measure the similarity of $\mathbf{v}^{}_i$ to a prototype it contributed to, which raises the similarity independently of whether $\ell^{}_i$ is correct. Excluding $\mathbf{v}^{}_i$ avoids this effect:
\begin{equation}
    m'_i = \cos\bigl(\mathbf{v}^{}_i, \boldsymbol{\mu}^{}_{\ell_i}\bigr) - \max_{c\, :\, \cos(\mathbf{t}^{}_c,\, \mathbf{t}^{}_{\ell_i}) < \tau} \cos\bigl(\mathbf{v}^{}_i, \boldsymbol{\mu}^{}_c\bigr),
\end{equation}
over the non-synonym classes instantiated in $\mathcal{M}$, converted at the same logit scale, $P^{\mathrm{coh}}_i = \sigma(\alpha\, m'_i)$; the signal is undefined when $\ell^{}_i$ has no other instance. The associated error is a labeling error, the same error the margin of \cref{eq:semantic} indicates: the two are alternative estimators rather than independent factors.
We therefore combine them as
\begin{equation}
    \bar{P}^{\mathrm{sem}}_i = \min\bigl(P^{\mathrm{sem}}_i,\, P^{\mathrm{coh}}_i\bigr),
    \label{eq:coherence}
\end{equation}
so that a low value from either estimator lowers the confidence. 
When $\ell^{}_i$ has no other instance in $\mathcal{M}$ and $P^{\mathrm{coh}}_i$ is undefined, $\bar{P}^{\mathrm{sem}}_i$ reduces to $P^{\mathrm{sem}}_i$.

\subsubsection{Vocabulary Membership}
A given object $o^{}_i$ whose true class $c$ is absent from $\mathcal{C}$ produces a near-uniform similarity profile, since the similarity between its embeddings $\mathbf{v}^{}_i$ and every class embedding in $\mathcal{C}$ is low. The semantic margin of \cref{eq:semantic} however measures only the difference between the top two similarities, and this difference can be large even when both similarities are low, resulting in assigning a label without evidence that the object belongs to $\mathcal{C}$.

We therefore compare the similarity of $\mathbf{v}^{}_i$ to $\mathcal{C}$ against its similarity to a separate set $\mathcal{N}$ of \emph{negative labels}~\cite{jiang2024negative}. $\mathcal{N}$ is constructed
by selecting words from a lexical database whose text embeddings are distant from every class in $\mathcal{C}$. We then define a membership probability, the proportion of total similarity assigned to $\mathcal{C}$ rather than to $\mathcal{N}$, computed at the same logit scale $\alpha$ used in \cref{eq:semantic}:
\begin{equation}
    P^{\mathrm{mem}}_i = \frac{\frac{1}{|\mathcal{C}|} \sum\limits_{c \in \mathcal{C}} e^{\alpha \cos(\mathbf{v}^{}_i,\, \mathbf{t}^{}_c)}}{\frac{1}{|\mathcal{C}|} \sum\limits_{c \in \mathcal{C}} e^{\alpha \cos(\mathbf{v}^{}_i,\, \mathbf{t}^{}_c)} + \frac{1}{|\mathcal{N}|} \sum\limits_{\mathbf{t} \in \mathcal{N}} e^{\alpha \cos(\mathbf{v}^{}_i,\, \mathbf{t})}} \,.
    \label{eq:membership}
\end{equation}
The above admits a Bayesian reading. Let the true class of $o_i$ lie in $\mathcal{C}$ or in $\mathcal{N}$ with equal prior probability and uniformly within each set, and let the likelihood of a candidate label be proportional to $e^{\alpha\cos(\mathbf{v}^{}_i,\, \cdot\,)}$.
Averaging that likelihood over the uniform stribution within each set gives the marginal likelihoods of $\mathcal{C}$ and $\mathcal{N}$, which are the two terms of the denominator of~\cref{eq:membership}; the numerator is the first of them.
With equal priors, $P^{\mathrm{mem}}_i$ is therefore the posterior probability that the true class of $o_i$ belongs to $\mathcal{C}$, and it does not depend on how many words each set contains.

When $o_i$ belongs to a class in $\mathcal{C}$, 
its similarity concentrates on that class and $P^{\mathrm{mem}}_i$ approaches one. 
When it does not, $\mathbf{v}^{}_i$ is no more similar to the classes of $\mathcal{C}$ than to the words of $\mathcal{N}$, and the posterior mass shifts towards $\mathcal{N}$.
The error this signal identifies is out-of-vocabulary content: an object entry whose true class does not belong to $\mathcal{C}$.

\subsubsection{Cross-view Consistency}
The preceding signals are evaluated over the fused embedding $\mathbf{v}^{}_i$: the fusion suppresses any disagreement among the per-view embeddings, and the signals inherit the resulting error~\cite{puigjaner2026relationship}. An object whose views are divided between two appearances is fused into the average of both. We recover the discarded disagreement from the fusion.

Let $\mathbf{r}^{}_i$ be the unweighted running mean of the $n$ per-view unit embeddings of $o^{}_i$. By the resultant-length identity for unit vectors, $\lVert \mathbf{r}^{}_i \rVert^2 = \tfrac{1}{n} + (1-\tfrac{1}{n})\,\bar{c}^{}_i$, where $\bar{c}^{}_i$ is the mean pairwise cosine similarity between the views. The resultant length depends on both quantities: for a fixed $\bar{c}^{}_i < 1$ it decreases as $n$ increases, so two objects of identical view agreement but different numbers of views would receive different values. We therefore recover $\bar{c}^{}_i$ from the identity, using the observed $\lVert \mathbf{r}^{}_i \rVert$ and $n$, and retain $\bar{c}^{}_i$ as the signal.

As a mean cosine similarity, $\bar{c}^{}_i \in [-1, 1]$, with $-1$ corresponding to complete disagreement, $1$ to complete agreement, and $0$ to views that are on average orthogonal. We rescale it to the unit interval by the unique affine map that fixes the two endpoints,
\begin{equation}
    P^{\mathrm{view}}_i = \frac{1 + \bar{c}^{}_i}{2},
    \label{eq:crossview}
\end{equation}
which equals one when every view produced the same embedding and decreases toward zero as the views disagree, independently of how many views were observed. The map is monotone and fixes the endpoints, so it preserves the ordering of $\bar{c}^{}_i$ and places the signal on the scale of the remaining factors. Unlike \cref{eq:semantic} and \cref{eq:membership}, it is not derived from a likelihood, and $P^{\mathrm{view}}_i$ is used as an estimate of the probability that the fusion represents a single object rather than as that probability itself.

A single view leaves $\bar{c}^{}_i$ undefined, and we set $P^{\mathrm{view}}_i = 1$. The convention is optimistic: an object observed once is not penalized on this factor, while the remaining three factors of \cref{eq:obj-prob} are defined for it and continue to constrain it. The quantity measures the agreement of the system's own observations, so it is paired with the union of the error types.

\subsection{Propagation through the Hierarchy}
\label{sec:propagation}

\subsubsection{Object-Level Probability}
\label{sec:object-probability}

Let $E^{}_s$ denote the event that the detection is real; $E^{}_{\mathrm{view}}$ that the fusion produced an embedding representing a single object; $E^{}_{\mathrm{mem}}$ that the object belongs to the vocabulary, and $E^{}_{\mathrm{sem}}$ that the assigned class is correct. An entry is reliable exactly when all four events hold jointly. The chain rule of probability factorizes $q^{}_i = P(E^{}_s, E^{}_{\mathrm{view}}, E^{}_{\mathrm{mem}}, E^{}_{\mathrm{sem}})$ exactly, and this factorization holds under any ordering of the four events. We order them as a detection first, its per-view observations fused second, and the vocabulary and label comparisons evaluated last, since this is the order in which the pipeline computes the underlying quantities $s^{}_i$, $P^{\mathrm{view}}_i$, $P^{\mathrm{mem}}_i$, and $\bar{P}^{\mathrm{sem}}_i$: each quantity is computed only from the evidence available up to its own stage, so each is a natural estimate of the probability of its event conditioned on the events computed before it.
The factorization reads

\begin{equation}
    \begin{aligned}
        q^{}_i ={}& P(E^{}_s) \\
        &\times P\bigl(E^{}_{\mathrm{view}} \mid E^{}_s\bigr) \\
        &\times P\bigl(E^{}_{\mathrm{mem}} \mid E^{}_s, E^{}_{\mathrm{view}}\bigr) \\
        &\times P\bigl(E^{}_{\mathrm{sem}} \mid E^{}_s, E^{}_{\mathrm{view}}, E^{}_{\mathrm{mem}}\bigr) .
    \end{aligned}
    \label{eq:obj-prob-chain}
\end{equation}
Each factor is then estimated by one stored quantity,
\begin{subequations}
\label{eq:obj-prob-subst}
\begin{align}
    s^{}_i &\approx P(E^{}_s), \label{eq:subst-s}\\
    P^{\mathrm{view}}_i &\approx P\bigl(E^{}_{\mathrm{view}} \mid E^{}_s\bigr), \label{eq:subst-view}\\
    P^{\mathrm{mem}}_i &\approx P\bigl(E^{}_{\mathrm{mem}} \mid E^{}_s, E^{}_{\mathrm{view}}\bigr), \label{eq:subst-mem}\\
    \bar{P}^{\mathrm{sem}}_i &\approx P\bigl(E^{}_{\mathrm{sem}} \mid E^{}_s, E^{}_{\mathrm{view}}, E^{}_{\mathrm{mem}}\bigr), \label{eq:subst-sem}
\end{align}
\end{subequations}
which gives
\begin{equation}
    q^{}_i \approx s^{}_i \, P^{\mathrm{view}}_i \, P^{\mathrm{mem}}_i \, \bar{P}^{\mathrm{sem}}_i .
    \label{eq:obj-prob}
\end{equation}

Each of $s^{}_i$, $P^{\mathrm{view}}_i$, and $P^{\mathrm{mem}}_i$ is computed independently of the others: $s^{}_i$ from the image region alone, $P^{\mathrm{view}}_i$ from the agreement between the per-view embeddings, and $P^{\mathrm{mem}}_i$ from the comparison of $\mathbf{v}^{}_i$ to $\mathcal{C}$ and $\mathcal{N}$ in \cref{eq:membership}. None of the three is computed by explicitly conditioning on the outcome of the events that precede it in \cref{eq:obj-prob-chain}, so using it as an estimate of the corresponding conditional probability is an approximation.
The substitution of \cref{eq:subst-sem} holds by construction. $P^{\mathrm{mem}}_i$ (\cref{eq:membership}) estimates the probability that the true class of $o^{}_i$ belongs to $\mathcal{C}$, while $\bar{P}^{\mathrm{sem}}_i$ (\cref{eq:coherence}) compares classes already known to lie in $\mathcal{C}$, so its value is only meaningful once vocabulary membership is established. Both quantities are computed from the same similarities between $\mathbf{v}^{}_i$ and $\{\mathbf{t}^{}_c\}$ (or, for the coherence term, $\{\boldsymbol{\mu}^{}_c\}$); multiplying their two marginal probabilities directly would count the evidence carried by $\mathbf{v}^{}_i$ twice. Conditioning $\bar{P}^{\mathrm{sem}}_i$ on $E^{}_{\mathrm{mem}}$, as \cref{eq:obj-prob-chain} does, avoids this double counting.

The four quantities $s^{}_i$, $P^{\mathrm{view}}_i$, $P^{\mathrm{mem}}_i$ and $\bar{P}^{\mathrm{sem}}_i$ are computed from observations of the same object $o^{}_i$, so any factor that degrades the quality of those observations tends to affect the four together rather than independently. The corresponding events are therefore positively associated, and for positively associated events the probability of the intersection is at least the product of the marginals~\cite{esary1967association},
\begin{equation}
    \begin{aligned}
    P\bigl(E^{}_s \cap &E^{}_{\mathrm{view}} \cap E^{}_{\mathrm{mem}} \cap E^{}_{\mathrm{sem}}\bigr) \\
    &\geq\; P(E^{}_s)\, P(E^{}_{\mathrm{view}})\, P(E^{}_{\mathrm{mem}})\, P(E^{}_{\mathrm{sem}}) .
    \end{aligned}
    \label{eq:obj-prob-bound}
\end{equation}
The quantities of \cref{eq:obj-prob-subst} are estimates, and the first three are computed without conditioning on the events that precede them, so \cref{eq:obj-prob} inherits the direction of \cref{eq:obj-prob-bound} without inheriting its guarantee. Positive association can only place $q^{}_i$ above a product of marginals, so \cref{eq:obj-prob} is a conservative estimate of $q^{}_i$ rather than a bound on it: the sign of the departure from independence is known, its magnitude and the error of the four estimates are not.

\subsubsection{Probability Propagation}
\label{sec:probability-propagation}

Let $\mathcal{O}(r) = \{o^{}_i : r(o^{}_i) = r\}$ be the objects assigned to room $r$. \Cref{eq:obj-prob} defines $q^{}_i$ as the probability that $o^{}_i$ is a correct instance of its assigned label $\ell^{}_i$. The same quantities define, for an arbitrary class $c \in \mathcal{C}$, the probability $q^{}_i(c)$ that $o^{}_i$ is an instance of $c$: the first three factors of \cref{eq:obj-prob} do not depend on the class and are kept, and the semantic factor is evaluated against $c$ in place of $\ell^{}_i$,
\begin{equation}
    \begin{aligned}
        m^{}_i(c) &= \cos(\mathbf{v}^{}_i, \mathbf{t}^{}_c) - \max^{}_{c' :\, \cos(\mathbf{t}^{}_{c'}, \mathbf{t}^{}_c) < \tau} \cos(\mathbf{v}^{}_i, \mathbf{t}^{}_{c'}),\\
        q^{}_i(c) &= s^{}_i\, P^{\mathrm{view}}_i\, P^{\mathrm{mem}}_i\, \bar{P}^{\mathrm{sem}}_i(c),
    \end{aligned}
    \label{eq:obj-prob-class}
\end{equation}
where $\bar{P}^{\mathrm{sem}}_i(c) = \min\bigl(\sigma(\alpha\, m^{}_i(c)),\, \sigma(\alpha\, m'_i(c))\bigr)$ is the semantic factor of \cref{eq:coherence}, with the coherence margin $m'_i(c)$ taken against the prototype $\boldsymbol{\mu}^{}_c$ of class $c$. When $\mathcal{M}$ holds no object labeled $c$ other than $o^{}_i$, no prototype exists and $\bar{P}^{\mathrm{sem}}_i(c) = \sigma(\alpha\, m^{}_i(c))$. By construction $q^{}_i(\ell^{}_i) = q^{}_i$. Since $\mathbf{t}^{}_c$ need not belong to the vocabulary used at construction, $c$ may be any phrase encoded at query time: the three class-independent factors and the similarities to $\{\mathbf{t}^{}_c\}_{c \in \mathcal{C}}$ are stored once per object, and $q^{}_i(c)$ then costs one text encoding and one cosine per object, the same as the similarity ranking of an open-vocabulary graph.

Room $r$ contains an instance of class $c$ if at least one $o^{}_i \in \mathcal{O}(r)$ is an instance of $c$. We evaluate this event over the objects the pipeline labeled $c$, $\mathcal{O}(r, c) = \{o^{}_i \in \mathcal{O}(r) : \ell^{}_i = c\}$, which removes the terms $q^{}_i(c)$ of objects labeled as a class other than $c$; the removed terms are largest for objects whose label is easily confused with $c$. Over $\mathcal{O}(r,c)$, and conditional on each $q^{}_i$ being an accurate per-object probability, the probability of the event is bracketed regardless of the correlation among objects as
\begin{equation}
    \max_{o^{}_i \in \mathcal{O}(r,c)} q^{}_i \;\leq\; P\big(y(r,c) = 1 \mid \mathcal{M}\big) \;\leq\; 1 - \prod_{o^{}_i \in \mathcal{O}(r,c)} \bigl(1 - q^{}_i\bigr).
    \label{eq:bracket}
\end{equation}
The lower bound holds unconditionally, since a single correctly labeled object is sufficient for $c$ to be present in $r$. The upper bound is the value under independence across objects (a noisy-OR), and it holds whenever the correctness events are positively associated~\cite{esary1967association}. Since objects are reconstructed based on the same detector, the same vocabulary, and from the same trajectory, a single error in any of these components mislabels different instances for the same underlying reason (\ie{} a class absent from $\mathcal{C}$, or one the model confuses with $c$, produces several objects labeled $c$ in one room). The upper bound considers their agreement as several independent detections of $c$, when it is one error repeated across objects, and it rises with every repetition. We therefore take the fully correlated limit, the lower bound, as the room-level belief,
\begin{equation}
    b(r, c) = \max^{}_{o^{}_i \in \mathcal{O}(r,c)} q^{}_i,
    \label{eq:room-belief}
\end{equation}
with $b(r,c) = 0$ when $\mathcal{O}(r,c)$ is empty. The belief is the probability that the most reliable object labeled $c$ in $r$ is a correct instance of it; further objects with the same label add no evidence, so a label repeated across a room by one shared error is not rewarded for being repeated.

\section{Experimental Evaluation}
\label{sec:experiments}

We evaluate the beliefs of \cref{eq:room-belief} on reconstructed graphs. \Cref{sec:setup} defines the dataset and the evaluation protocol. \Cref{sec:search} measures the belief on an object search task. 
\Cref{sec:room_belief} evaluates the propagated room-level beliefs against the graph as built and against the aggregation under independence.

\subsection{Dataset and Protocol}
\label{sec:setup}

We adopt the Habitat-Matterport~3D Semantics (HM3DSem) dataset~\cite{yadav2023hm3dsem} and select six scenes following the protocol in~\cite{werby2024hierarchical}. We build one 3DSG per scene and evaluate it on the reconstructed graphs, so every quantity entering \cref{eq:obj-prob} is one the system itself produced. 
A reconstructed object is associated with an annotated instance by spatial proximity, which gives its true class, and the presence indicator $y(r,c)$ is one when an annotated instance of class $c$ lies inside the extent of room $r$.

We define an \emph{entry} as a pair $(r,c)$ for which the graph holds at least one object of class $c$ in room $r$, that is, for which the set $\mathcal{O}(r,c)$ of \cref{eq:room-belief} is non-empty; $b(r,c)$ is its belief. An entry asserts that room $r$ contains an object of class $c$, and this assertion is \emph{correct} when $y(r,c) = 1$ and \emph{incorrect} otherwise. A pair with $\mathcal{O}(r,c) = \emptyset$ is one for which the graph holds no object of class $c$ in room $r$, and its belief is zero under every configuration, since \cref{eq:room-belief} is taken over $\mathcal{O}(r,c)$. For a pair whose class is present but was not recovered by the construction pipeline, $y(r,c) = 1$ while the graph holds no entry, and no configuration of the belief distinguishes it from a pair whose class is absent.

A \emph{query} is a pair of a scene and a class for which the graph holds at least one object of that class anywhere in the scene. The queries are not sampled from $\mathcal{C}$: every class the pipeline instantiated is issued, including those that no annotated instance of the scene confirms, and a class instantiated in several scenes is queried once in each. The six evaluated scenes instantiate between $72$ and $100$ distinct classes each, for $500$ scene--class queries in total. The remaining categories of $\mathcal{C}$ are those the assignment of \cref{sec:scene-graph} selected for no object of any scene, and they enter no query. A query and an entry apply one condition at two scales, the scene and the room, so every query has at least one candidate room.

The vocabulary $\mathcal{C}$ is the HM3DSem category inventory, defined independently of the reconstructed graphs, and the negative labels $\mathcal{N}$ of \cref{eq:membership} are mined once from a lexical database~\cite{mccrae2019english} relative to $\mathcal{C}$ and reused across scenes. The system is run with $\mathcal{C}$ as its vocabulary, so every object of the graph carries a class from one set.

The framework requires only the interface of \cref{sec:scene-graph} and applies to any 3DSG that stores those quantities. We instantiate it on HOV-SG~\cite{werby2024hierarchical}. The system itself is not modified and builds its graph as published with the belief exploiting the quantities that the graph already stores. We evaluate the system the way it builds the graph and with the beliefs of \cref{eq:room-belief} added.

\subsection{Object Search}
\label{sec:search}

\begin{table}[t]
    \centering
    \caption{\textbf{Object-search performance.} Per-scene results compare the baseline HOV-SG~\cite{werby2024hierarchical} ranking with the query-conditioned object belief where OS@$1$--$5$ are reported in percent. Higher is better. 
    }
    \label{tab:object_search_results}
    \small
    \resizebox{\columnwidth}{!}{
    \begin{tabular}{llcccc}
        \toprule
        Scene
        & Method
        & OS@$1$
        & OS@$3$
        & OS@$5$
        & OMRR@$5$ \\
        \midrule

        \multirow{2}{*}{00824}
        & HOV-SG
        & $\mathbf{28.00}$ & $\mathbf{30.67}$ & $\mathbf{30.67}$ & $\mathbf{0.29}$ \\
        & {Belief} (\emph{ours})
        & $26.67$ & $\mathbf{30.67}$ & $\mathbf{30.67}$ & $0.28$ \\

        \midrule

        \multirow{2}{*}{00829}
        & HOV-SG
        & $22.22$ & $26.39$ & $27.78$ & $0.25$ \\
        & {Belief} (\emph{ours})
        & $\mathbf{29.17}$ & $\mathbf{30.56}$ & $\mathbf{30.56}$ & $\mathbf{0.30}$ \\

        \midrule

        \multirow{2}{*}{00843}
        & HOV-SG
        & $15.00$ & $16.25$ & $17.50$ & $0.16$ \\
        & {Belief} (\emph{ours})
        & $\mathbf{16.25}$ & $\mathbf{18.75}$ & $\mathbf{18.75}$ & $\mathbf{0.17}$ \\

        \midrule

        \multirow{2}{*}{00847}
        & HOV-SG
        & $22.99$ & $\mathbf{31.03}$ & $\mathbf{31.03}$ & $0.26$ \\
        & {Belief} (\emph{ours})
        & $\mathbf{26.44}$ & $\mathbf{31.03}$ & $\mathbf{31.03}$ & $\mathbf{0.29}$ \\

        \midrule

        \multirow{2}{*}{00849}
        & HOV-SG
        & $\mathbf{30.23}$ & $\mathbf{34.88}$ & $34.88$ & $\mathbf{0.32}$ \\
        & {Belief} (\emph{ours})
        & $26.74$ & $33.72$ & $\mathbf{36.05}$ & $0.31$ \\

        \midrule

        \multirow{2}{*}{00877}
        & HOV-SG
        & $26.00$ & $35.00$ & $37.00$ & $0.30$ \\
        & {Belief} (\emph{ours})
        & $\mathbf{32.00}$ & $\mathbf{37.00}$ & $\mathbf{38.00}$ & $\mathbf{0.34}$ \\

        \midrule

        \multirow{2}{*}{{Macro}}
        & HOV-SG
        & $24.07$ & $29.04$ & $29.81$ & $0.26$ \\
        & {Belief} (\emph{ours})
        & $\mathbf{26.21}$ & $\mathbf{30.29}$ & $\mathbf{30.84}$ & $\mathbf{0.28}$ \\
        \bottomrule
    \end{tabular}
    }
\end{table}

Open-vocabulary 3DSGs answer a language query by retrieving the objects of the scene that match it. We adopt the object retrieval protocol of HOV-SG: every class the pipeline instantiated in a scene is issued as a query, and the graph returns the objects that best match it. HOV-SG grounds the query in the similarity between the query and the object embeddings. Our method replaces this grounding with the query-conditioned probability $q^{}_i(c)$ of \cref{eq:obj-prob-class}, so the two are evaluated on identical graphs and queries and differ only in how the query is grounded.

\subsubsection{Metrics}
\label{sec:search_metrics}

A retrieval is successful when the returned object is associated with an annotated instance of the queried class (\cref{sec:setup}). We report the \textbf{Success Rate} within the top $k$ retrieved objects (OS@$k$, $k \in \{1,3,5\}$) and the mean reciprocal rank of the first successful retrieval within the top five (OMRR@5), per scene and averaged over scenes.

\subsubsection{Results}
\label{sec:search_results}

\Cref{tab:object_search_results} reports the retrieval results over the evaluated scenes. The gain is concentrated at the first rank. OS@$5$ rises by $1.03$ against $2.14$ at OS@$1$, so at most $1.03$ of the first-rank gain can be attributed to objects that the similarity grounding did not return within five ranks at all, and the remaining $1.11$ is obtained by reordering objects it had already returned. The same effect closes the interval between the two ranks: OS@$5$ exceeds OS@$1$ by $5.74$ under the similarity grounding and by $4.63$ under the probability. The two groundings, therefore, return largely the same five objects, and our belief ranks them correctly.
 
The gain is not uniform across scenes. OS@$1$ increases in four of the six, by $6.95$ points in scene 00829 and by $6.00$ in 00877, and decreases in the two scenes in which the similarity grounding is already strongest, by $1.33$ points in 00824 and by $3.49$ in scene 00849. In 00824 the two groundings achieve identical OS@$3$ and OS@$5$, so the loss is a reordering within one retrieved set; in 00849 OS@$5$ increases by $1.17$ while OS@$1$ decreases, so the probability returns an object the similarity had missed but at the same time demotes one the similarity had ranked first correctly.

\subsection{Room-Level Belief}
\label{sec:room_belief}

\begin{table}[t]
    \centering
    \caption{\textbf{Room-level belief quality, by class frequency} on HM3DSem~\cite{yadav2023hm3dsem} over object-supported room--class pairs. Values report the Brier score and AUROC. All: averaged over the 6 evaluated scenes. \emph{Frequent}, \emph{Common}, \emph{Rare} pool pairs across evaluated scenes: 6 or more, 3--5, and 1--2 scenes, respectively. The 95\% intervals use 10,000 scene-bootstrap resamples and are reported only for macro Brier. Bold marks the best value per column.}
    \label{tab:room_belief}
    \footnotesize
    \setlength{\tabcolsep}{4pt}
    \begin{tabular}{lccccc}
        \toprule
        & Brier $\downarrow$ & AUROC $\uparrow$ & \multicolumn{3}{c}{Brier $\downarrow$} \\
        \cmidrule(lr){2-2} \cmidrule(lr){3-3} \cmidrule(lr){4-6}
        Method & All & All & Frequent & Common & Rare \\
        \midrule
        As built & $0.80$ & $0.50$ & $\mathbf{0.17}$ & $0.74$ & $0.91$ \\
        & \ci{0.76,\,0.83} & & & & \\
        Noisy-OR & $0.47$ & $0.54$ & $0.23$ & $0.44$ & $0.49$ \\
        & \ci{0.44,\,0.50} & & & & \\
        {Belief} (\emph{ours}) & $\mathbf{0.33}$ & $\mathbf{0.61}$ & $0.25$ & $\mathbf{0.32}$ & $\mathbf{0.33}$ \\
        & \ci{0.30,\,0.34} & & & & \\
        \bottomrule
    \end{tabular}
    \par\vspace{1mm}
\end{table}

\begin{table}[t]
    \centering
    \caption{\textbf{Room-level belief quality} by scene on HM3DSem~\cite{yadav2023hm3dsem}. Values report the Brier score (lower is better). Best results per scene are shown in bold.}
    \label{tab:room_belief_scenes}
    \scriptsize
    \setlength{\tabcolsep}{5pt}
    \begin{tabular*}{\columnwidth}{@{\extracolsep{\fill}}lcccccc@{}}
        \toprule
        Method & 00824 & 00829 & 00843 & 00847 & 00849 & 00877 \\
        \midrule
        As built
        & $0.80$ & $0.76$ & $0.87$ & $0.82$ & $0.77$ & $0.74$ \\
        Noisy-OR
        & $0.49$ & $0.43$ & $0.52$ & $0.50$ & $0.43$ & $0.46$ \\
        {Belief} (\emph{ours})
        & $\mathbf{0.33}$ & $\mathbf{0.31}$ & $\mathbf{0.35}$ & $\mathbf{0.36}$ & $\mathbf{0.28}$ & $\mathbf{0.33}$ \\
        \bottomrule
    \end{tabular*}
\end{table}

The search of \cref{sec:search} evaluated the beliefs of one class at a time. Here we evaluate the propagated beliefs at a room level over all classes: for a room $r$, whether $b(r,c)$ is high for the classes present in $r$ and low for those absent.

The population is the set of entries of \cref{sec:setup}: the room--class pairs $(r,c)$ for which the graph holds at least one object of class $c$ in $r$.
A class present in a room without an entry is a miss of the construction pipeline, which no belief reaches.

\subsubsection{Baselines}
\label{sec:room_baselines}

We define two baselines that read the same graph as the belief and differ only in the value attached to an entry. \textbf{As built} is the graph as the system emits it: every entry is asserted with certainty, $b(r,c) = 1$. \textbf{Noisy-OR} is the upper bound of \cref{eq:bracket}, the aggregation under independence across objects.

\subsubsection{Metrics}
\label{sec:room_belief_metrics}

We evaluate each entry with the \textbf{Brier score}~\cite{gneiting2007strictly}, the squared difference $\bigl(b(r,c) - y(r,c)\bigr)^2$ between the value assigned to the entry and the presence indicator.

We report it under two aggregations. The first averages the entries of a room, then the rooms of a scene, then the scenes, with a bootstrap $95\%$ interval over scenes, so that each room contributes once irrespective of the number of classes it holds (\emph{All}). The second pools all entries and partitions them by the number of annotated scenes in which the class of the entry is present, into \emph{Frequent} (six or more), \emph{Common} (three to five) and \emph{Rare} (one or two), with one score per bin; the entries whose class is present in no annotated scene enter \emph{All} alone.

The Brier score rewards a value that is both correctly ordered and correctly scaled. To isolate the first, we also report the area under the ROC curve (\textbf{AUROC}) of each value over the entries of a scene, correct against incorrect, averaged over scenes: the probability that a correct entry 
scores above an incorrect one, 
ties counted as one half.
Being invariant to any monotone rescaling, it measures ordering alone.

\subsubsection{Results}
\label{sec:room_belief_results}

\Cref{tab:room_belief} reports the aggregate results and \cref{tab:room_belief_scenes} the decomposition by scene.
 
The graph \emph{as built} asserts every entry with certainty and achieves a Brier score of $0.80$, which under this averaging is the fraction of its entries that are incorrect: close to four entries in five assert a class the room does not contain. Its AUROC is $0.50$, since a constant value admits no ordering.
 
Our \emph{belief} lowers the Brier score to $0.33$, with a bootstrap interval of $[0.30, 0.34]$ disjoint from the interval $[0.76, 0.83]$ of the graph \emph{as built}, and raises the AUROC to $0.61$. The reduction holds in each of the six scenes, from $0.87$ to $0.35$ in scene 00843 and from $0.74$ to $0.33$ in scene 00877, and the range of the belief across scenes, $0.28$ to $0.36$, is narrower than that of the graph \emph{as built}, $0.74$ to $0.87$. The \emph{belief} therefore separates correct from incorrect entries at a rate above chance, and the value it assigns is closer to the observed correctness rate than the certainty the graph emits.
 
Compared to the aggregation under independence of the \emph{Noisy-OR}, our method is lower in each of the six scenes, and the two intervals are disjoint. Its direction is the one predicted in \cref{sec:probability-propagation}: the objects of a room are labeled by one detector, against one vocabulary, from one trajectory, so the noisy-OR raises the belief with every repetition of an error that is not independent of the repetitions preceding it, and the incorrect entries are the entries it raises most.
 
The three frequency bins separate the regimes in which each method is accurate. On \emph{frequent} classes the entries of the graph are predominantly correct, and the graph as built achieves the lowest Brier score of the three methods, $0.17$, ahead of the noisy-OR at $0.23$ and of the belief at $0.25$. On rare classes the ordering is reversed: the graph \emph{as built} attains $0.91$, the noisy-OR $0.49$ and the belief $0.33$. The belief is the only method whose score varies little across the three bins, from $0.25$ to $0.33$, a range of $0.08$ against $0.26$ for the noisy-OR, and $0.74$ for the graph as built. Its error is close to uniform in the frequency of the queried class, whereas every other method has a regime in which its error is several times its aggregate value.
\section{Conclusion}

Open-vocabulary 3DSGs compute a detector confidence and a set of language-aligned embeddings for every object they store, and discard both at query time, so every entry of the graph is read as certain. We defined four per-object signals over these stored quantities, converted them to probabilities and propagated the result along the containment edges into a belief that a room contains a queried class.

We show on HM3DSem that grounding a language query in the per-object probability rather than in the raw similarity improves the retrieval of objects, and that the belief propagated from the same probabilities lowers the error of the room-level assertions of the graph it reads.

The defined belief encodes no knowledge of what a room of a given type usually contains. In future work, we plan on combining our belief with a scene-independent prior over room types into a single posterior, since the two are functions of disjoint evidence and we expect them to be accurate in complementary regimes, and extending the propagation to the uncertainty of the containment edge itself.
\bibliographystyle{IEEEtran}
\bibliography{bibliography}

\end{document}